\documentclass[11pt]{article}

\usepackage[preprint]{acl}

\usepackage{times}
\usepackage{latexsym}
\usepackage[T1]{fontenc}
\usepackage[utf8]{inputenc}
\usepackage{microtype}
\usepackage{inconsolata}
\usepackage{graphicx}
\usepackage{booktabs}
\usepackage{enumitem}
\usepackage{amssymb}
\usepackage{siunitx}
\usepackage{amsmath}
\usepackage{multirow}
\usepackage{enumitem}
\usepackage{xcolor}
\usepackage{pifont}

\newcommand{\cmark}{\textcolor{green!60!black}{\ding{51}}}
\newcommand{\xmark}{\textcolor{red!70!black}{\ding{55}}}

\title{Different Perturbations, Different Mechanisms: Understanding Continued Pre-training for Zero-Shot Dialect Robustness}

\author{Aarohi Srivastava \and David Chiang \\
    Computer Science and Engineering \\
 University of Notre Dame \\
 Notre Dame, IN, USA \\ \texttt{\{asrivas2, dchiang\} @nd.edu}}

\begin{document}
\maketitle

\begin{abstract}
Dialectal variation remains a major challenge for multilingual language models. Perturbation-based continued pre-training (CPT) has emerged as a promising approach to improving robustness, yet existing work largely evaluates individual perturbation strategies in isolation and provides limited insight into why they work. We present a systematic study of perturbation-based CPT for multilingual dialect robustness in LLMs, comparing six training conditions across nine German, Italian, and Arabic dialect tasks. Perturbation-based CPT, especially character-noised CPT, consistently improves zero-shot dialect robustness while largely preserving standard variety performance. More importantly, we show that methods with similar downstream performance induce distinct mechanisms of robustness, exhibiting different patterns of language model adaptation, representational alignment, and prediction repair. Our results provide a more complete understanding of how synthetic surface variation improves robustness and offer practical guidance for selecting CPT strategies in multilingual and dialectal settings.
\end{abstract}

\section{Introduction}

Large language models (LLMs) are trained primarily on standardized, high-resource text, yet real-world language use is inherently diverse. Differences in spelling, lexical choice, morphology, and orthographic conventions across dialects can substantially degrade downstream NLP performance, even when the underlying meaning remains clear to human readers \citep{kantharuban2023quantifying,faisal-etal-2024-dialectbench}. As multilingual LLMs are increasingly deployed in real-world settings, improving robustness to naturally occurring linguistic variation has become an important challenge.

A promising approach is perturbation-based continued pre-training (CPT), which exposes models to synthetic variations of standard variety text without requiring dialect-specific supervision. Existing work has shown that such perturbations can improve robustness, but largely evaluates them through downstream benchmark performance. As a result, we know comparatively little about how different perturbation strategies improve robustness, whether they induce similar representations, and when one form of synthetic variation should be preferred over another.

We present a systematic study of perturbation-based CPT for multilingual dialect robustness. We compare six continued pre-training strategies spanning character-level noise, subword regularization, subword perturbations, and phonologically motivated transformations across German, Italian, and Arabic. Rather than evaluating perturbations solely by downstream accuracy, we complement benchmark results with analyses of language model adaptation, representation similarity, tokenization behavior, and prediction repair to understand the mechanisms through which robustness emerges.

Our experiments show that perturbation-based CPT consistently improves zero-shot dialect robustness while largely preserving performance on standard varieties. Character-level noise emerges as the strongest general-purpose strategy, but each method builds robustness by interacting with different mechanisms of the underlying model. Random and phonology-based character perturbations are associated with increased alignment between standard and dialect representations, while clean and token-level perturbations are associated more strongly with improvements in language model fit. These findings suggest that perturbation-based CPT is best understood not as a single robustness intervention, but as a family of adaptation strategies that encourage different forms of linguistic invariance.

Our contributions are as follows:
\begin{enumerate}[itemsep=0pt, parsep=0pt]
    \item We present a systematic comparison of multiple perturbation-based CPT strategies for multilingual dialect robustness across three languages, nine downstream tasks, and a diverse collection of dialect varieties.
    
    \item We show that perturbation-based CPT is a reliable robustness intervention, improving 43 of 45 language--task--method comparisons while largely preserving standard variety performance. Character-level perturbations provide the strongest general-purpose strategy, although no method is universally optimal.
    
    \item We demonstrate that perturbation strategies with similar downstream performance can induce distinct mechanisms of robustness. Character-level perturbations are associated more strongly with representational alignment, while clean and token-level perturbations are associated more strongly with improvements in language model fit.
\end{enumerate}

\section{Background and Related Work}

\paragraph{Robustness to Dialectal Variation}

Natural language varieties remain a persistent challenge for NLP systems. Even relatively small differences in spelling, lexical choice, morphology, and orthographic convention can substantially alter tokenization and downstream representations, leading to degraded performance across tasks and languages \citep{kumar2020noisy, soper2021bart, yin2020robustness, aepli2022improving, kantharuban2023quantifying, faisal-etal-2024-dialectbench}. Recent benchmark studies show that these robustness gaps persist in multilingual and instruction-tuned LLMs \citep{faisal-anastasopoulos-2025-testing}, motivating methods that generalize beyond standardized language.

\paragraph{Dialect Adaptation}

Most approaches improve dialect robustness by adapting models to target varieties through architectural modifications or dialect-specific training. Prior work has explored vocabulary adaptation, alternative tokenization schemes, character-aware models, and adapter-based approaches that encourage dialect-invariant representations \citep{pfeiffer2020mad, lee2021korean, charformer, wang2021multi, held2023tada, han2025adapters}. More recently, decoder-only LLMs have been adapted using LoRA-based dialect adapters trained on dialect resources \cite{srirag2025predicting, maheshwari2026improving}, while broader adaptation pipelines have highlighted continued pre-training on general dialect text as an important component of dialect adaptation \citep{painter2026diallm}. Although effective, these methods typically rely on target-dialect supervision or dialect-specific resources. In contrast, we investigate whether robustness can emerge from synthetic surface variation alone, enabling transfer to unseen dialects without dialect supervision.

\paragraph{Synthetic Surface Variation}

Synthetic perturbations have long been used to improve robustness to noisy and nonstandard text. Character-level perturbations improve robustness across machine translation and language understanding tasks \citep{karpukhin2019training, vaibhav2019improving, yin2020robustness}, and have subsequently been applied to zero-shot dialect robustness through perturbation-based continued pre-training in encoder models \citep{aepli2022improving, srivastava2023fine, blaschke2023does, srivastava2023bertwich}. More recently, \citet{kojima2025continual} demonstrated that noisy continued pre-training similarly improves robustness in decoder-only LLMs for English text correction tasks, and \citet{bafna2025dialup} utilized linguistically-motivated perturbations for dialectal machine translation.

Despite these advances, existing work largely evaluates individual perturbation strategies in isolation and primarily through downstream benchmark performance. Consequently, it remains unclear whether different forms of synthetic surface variation induce the same kind of robustness or improve performance through different mechanisms, limiting guidance for both researchers and practitioners. Our work addresses this gap through a controlled comparison of multiple perturbation-based CPT strategies, complemented by analyses of language model adaptation, representational alignment, tokenization behavior, and prediction repair.

\section{Method}

\begin{table}[]
\centering
\small
\begin{tabular}{@{}l*{5}{l}@{}}
\toprule
\textbf{Variation} & \multicolumn{5}{l}{\textbf{Token Sequence}} \\ \midrule
\texttt{Clean}/Standard & \_Reg & net & \_es & \_heute &        \\
\texttt{BPE-Drop}       & \_Reg & net & \_es & \_he    & ute    \\
\texttt{Sub-Rep}        & \_Reg & zu  & \_es & \_heute &        \\
\texttt{Noisy}          & \_Reg & net & \_es & \_hel   & te     \\
\texttt{Phon}          & \_Reg & net & \_ec & \_hö   & uta     \\
Dialect        & \_Reg & nt  & s    & \_he    & inte   \\
\bottomrule
\end{tabular}
\caption{\textbf{Illustrative examples of synthetic perturbations, dialectal variation, and resulting tokenizations.} The clean sentence is drawn from a German intent detection dataset, and the dialect example is its parallel South Tyrolean variant. All examples use the Llama-2-7B tokenizer. Underscores denote word-initial tokens following Llama tokenization conventions.}
\label{tab:examples}
\end{table}

Recent work has identified CPT as an important stage of dialect adaptation pipelines in LLMs \citep{painter2026diallm}. Our goal is to understand how different forms of synthetic surface variation influence zero-shot robustness to dialectal language. We compare a family of CPT methods that expose the model to different forms of synthetic variation while holding all other aspects of training constant. 

\paragraph{Training} We continue pre-training Llama-2-7B\footnote{\url{https://huggingface.co/meta-llama/Llama-2-7b}} on the German, Italian, and Arabic portions of multilingual C4 \cite{2020t5}. Using a single model architecture, corpus, and training procedure isolates the effect of perturbation strategy from architectural and training differences.  Following \citet{kojima2025continual}, our CPT is performed using parameter-efficient fine-tuning with LoRA \citep{hu2021lora}; training hyperparameters and perturbation calibration values are provided in Table~\ref{tab:calibration}. Because mC4 primarily contains standard variety web text, any dialectal content is incidental rather than curated, allowing synthetic perturbations to serve as the primary source of robustness-inducing variation. We find the main trends on a second 7B model family (Qwen2-7B). The complete training configuration is reported in Appendix~\ref{sec:hyperparams}.

Training loss consistently plateaued after approximately 2500 optimization steps. To avoid implicit tuning on downstream evaluation tasks, we do not select checkpoints based on downstream performance; instead, all reported results use the 2500-step checkpoint. Unless otherwise noted, results are averaged across three independent CPT runs with different random seeds, and we report the mean and standard deviation across runs.

\subsection{CPT Modes}

We compare six training conditions: \textsc{Base} (no CPT), \textsc{Clean} (continued pre-training on unmodified text), and four perturbation-based variants that modify different aspects of the input. Following \citet{kojima2025continual}, perturbation-based methods leave each document unchanged with probability $0.5$; otherwise, a document-specific perturbation rate is sampled uniformly from a predefined interval. This exposes the model to a continuum of perturbation strengths while isolating the effect of the perturbation itself. Table~\ref{tab:examples} illustrates the resulting text and tokenization changes.

\paragraph{Clean CPT (\textsc{Clean}).}
\textsc{Clean} performs CPT on unmodified mC4 and controls for additional exposure to the target language without perturbation.

\paragraph{Character-Noised CPT (\textsc{Char}).}
Eligible words undergo at most one character insertion, deletion, substitution, or adjacent-character transposition using characters from the target language alphabet \cite{srivastava2023bertwich, kojima2025continual}. Perturbations modify both token identity and subword segmentation without attempting to mimic any particular dialect, encouraging robustness to naturally occurring surface variation.

\paragraph{BPE Dropout CPT (\textsc{BPE-Drop}).}
BPE dropout stochastically suppresses tokenizer merge operations, exposing the model to alternative subword segmentations while preserving the underlying character sequence. Maximum dropout probabilities are calibrated separately for each language to match the tokenization changes induced by \textsc{Char}.

\paragraph{Subword Replacement CPT (\textsc{Sub-Rep}).}
Input tokens are replaced with vocabulary items sampled uniformly from the tokenizer vocabulary while preserving segmentation boundaries. Special tokens, punctuation, and numeric tokens are excluded. Because random token replacement introduces substantially greater semantic distortion than character-level perturbations, we use a maximum replacement probability of 5\% for all languages.

\paragraph{Phonologically-Noised CPT (\textsc{Phon}).}
We apply the phonological noiser from DialUp \citep{bafna2024evaluating,bafna2025dialup}, which models systematic phonological correspondences between standard language and dialect while preserving lexical content. Unlike the lexical and morphological components of DialUp, the phonological noiser operates directly on surface forms without requiring learned language-specific models or dialect resources, making it directly comparable to the other perturbation strategies. Maximum perturbation rates are calibrated separately for each language to match the tokenization changes induced by \textsc{Char}.

\subsection{Matching Perturbation Strength}

To compare perturbation \emph{types} rather than perturbation \emph{severity}, we calibrate perturbation strategies to induce comparable tokenization changes. We first measure the average token inflation and characters per token produced by \textsc{Char} on multilingual C4 samples, then select the maximum perturbation parameters for \textsc{BPE-Drop} and \textsc{Phon} to match these statistics. Because \textsc{Sub-Rep} alters token identities rather than segmentation boundaries, directly matching tokenization statistics is not meaningful. Instead, we select a replacement rate that produces a comparable overall level of perturbation to \textsc{Char}. Perturbation rates are sampled uniformly over predefined intervals during continued pre-training. This exposes the model to a range of perturbation strengths while maintaining comparable average perturbation severity across methods. Calibrated probabilities are reported in Table~\ref{tab:calibration}.

\subsection{Tokenization Comparison}

The perturbation strategies manipulate different aspects of subword structure while remaining comparable in strength. Thus, before evaluating downstream performance, we examine whether they produce tokenization changes similar to those observed in naturally occurring dialect text.

Table~\ref{tab:examples} provides qualitative examples of how dialectal spellings and synthetic perturbations alter subword structure. We further compare measures such as average subword tokens per word and average characters per token (Table~\ref{tab:tok_stats}). Synthetic perturbations broadly reproduce the tokenization changes observed in dialectal text, supporting their use as controlled robustness-inducing interventions in the downstream experiments.

\begin{table}[]
\setlength{\tabcolsep}{1.8pt}
\centering
\small
\begin{tabular}{@{}llrrr@{}}
\toprule
\textbf{Lang.}    & \textbf{} & \textbf{Tok/Word} & \textbf{Char/Tok} & \textbf{\%Words $\geq$3 Tok} \\ \midrule
\textbf{DE}  & Standard  & 1.78              & 3.29              & 49.04                       \\
                 & Dialect   & 1.88              & 2.82              & 57.47                       \\
                 & Char Noise     & 1.96              & 2.98              & 56.22                       \\
\textbf{IT} & Standard  & 1.82              & 3.12              & 51.33                       \\
                 & Dialect   & 2.14              & 2.58              & 74.95                       \\
                 & Char Noise     & 1.98              & 2.87              & 57.25                       \\
\textbf{AR}  & Standard  & 5.43              & 1.10              & 99.19                       \\
                 & Dialect   & 5.46              & 1.10              & 98.55                       \\
                 & Char Noise     & 5.48              & 1.09              & 98.41                       \\ \bottomrule
\end{tabular}
\caption{\textbf{Tokenization Statistics.} Results are computed on parallel standard–dialect examples and averaged across dialects within each language. \emph{Char Noise} applies a fixed 25\% word-level corruption rate. \textit{Dialect} and \textit{Char Noise} text exhibit similar increases in subword fragmentation relative to \textit{standard} text.}
\label{tab:tok_stats}
\end{table}

\section{Languages, Tasks, and Evaluation}

\begin{table*}
\centering
\footnotesize
\setlength{\tabcolsep}{2pt}
\begin{tabular}{ll|cccccc}
\toprule
&& \multicolumn{6}{c}{\textbf{Dialect Average}} \\
\textbf{Lang} & \textbf{Task} & \textbf{\textsc{Base}} & \textbf{\textsc{Clean}} & \textbf{\textsc{BPE-Drop}} & \textbf{\textsc{Sub-Rep}} & \textbf{\textsc{Char}} & \textbf{\textsc{Phon}} \\
\midrule
\textbf{German} & Intent & 61.1 & 65.4 $\pm$ 1.9 & 63.3 $\pm$ 1.4 & 64.1 $\pm$ 1.7 & \textbf{71.2 $\pm$ 0.8} & 67.4 $\pm$ 0.2 \\
\textbf{German} & Slot & 14.7 & 16.1 $\pm$ 0.6 & 15.7 $\pm$ 0.3 & 17.0 $\pm$ 0.8 & \textbf{18.7 $\pm$ 0.5} & 16.9 $\pm$ 0.3 \\
\textbf{German} & NER & 42.7 & 42.7 $\pm$ 1.4 & 43.3 $\pm$ 0.5 & 43.0 $\pm$ 1.5 & \textbf{45.7 $\pm$ 0.1} & 45.1 $\pm$ 0.2 \\
\midrule
\textbf{Italian} & Intent & 70.4 & 74.1 $\pm$ 0.6 & 72.7 $\pm$ 2.3 & 74.0 $\pm$ 1.0 & \textbf{76.5 $\pm$ 2.2} & 75.5 $\pm$ 0.7 \\
\textbf{Italian} & Slot & 18.1 & 20.2 $\pm$ 1.5 & 20.1 $\pm$ 0.9 & \textbf{21.1 $\pm$ 1.3} & 18.9 $\pm$ 0.3 & 19.5 $\pm$ 0.5 \\
\textbf{Italian} & NER & 51.6 & 54.5 $\pm$ 1.1 & 53.8 $\pm$ 1.6 & 54.8 $\pm$ 0.8 & \textbf{55.6 $\pm$ 1.5} & 51.6 $\pm$ 0.4 \\
\midrule
\textbf{Arabic} & Topic & 73.9 & 76.8 $\pm$ 0.6 & 76.1 $\pm$ 0.6 & 76.5 $\pm$ 0.8 & \textbf{77.6 $\pm$ 0.6} & 76.4 $\pm$ 0.1 \\
\textbf{Arabic} & Sentiment & 54.5 & 59.1 $\pm$ 1.0 & \textbf{59.4 $\pm$ 0.8} & 58.7 $\pm$ 0.6 & 59.3 $\pm$ 1.3 & 59.0 $\pm$ 1.0 \\
\textbf{Arabic} & NER & 50.3 & \textbf{54.0 $\pm$ 1.2} & 52.9 $\pm$ 1.1 & 53.5 $\pm$ 0.7 & 52.7 $\pm$ 1.3 & 50.1 $\pm$ 0.1 \\
\bottomrule
\end{tabular}

\caption{\textbf{Dialect-average performance across tasks and languages.} For each CPT condition, we report the mean and standard deviation across three random seeds; \textsc{Base} is evaluated once and has no standard deviation. Dialect averages are computed in two stages: we first average across dialect varieties within each seed, then compute the mean and standard deviation across the three seed-level averages. Best (highest) results within each row are in bold. Analogous results for standard varieties and individual varieties are reported in Table~\ref{tab:full}.}
\label{tab:combined_main}
\end{table*}

\paragraph{Languages} We evaluate robustness across German, Italian, and Arabic not only because they are typologically distinct languages, but also because their dialects represent different forms of language variation. In the datasets considered, German varieties differ largely through spelling variation, Italian varieties exhibit greater regional lexical variation alongside nonstandard orthography, while Arabic comprises geographically distributed regional varieties with broader lexical, morphological, and orthographic divergence from Modern Standard Arabic (MSA). These languages provide complementary settings for evaluating how different forms of synthetic surface variation can build robustness across different types of linguistic variation.

\paragraph{Tasks} Evaluation spans sequence- and token-level prediction tasks. For all downstream tasks, classifiers are trained exclusively on standard variety training data and evaluated on both standard and dialect test sets. This protocol measures zero-shot robustness to natural linguistic variation rather than the ability to adapt through dialect-specific supervision. Table~\ref{tab:tasks} summarizes the downstream tasks and evaluated language varieties, and Table~\ref{tab:dataset_stats} reports dataset statistics.

The evaluation suite includes intent detection, slot filling, named entity recognition (NER), sentiment analysis, and topic identification. Dialect categorization for all Arabic tasks, as well as German and Italian NER, follows DialectBench~\citep{faisal-etal-2024-dialectbench}. Intent detection and slot filling are based on xSID4LR~\citep{van-der-goot-etal-2021-masked}, with dialect categorization following the corresponding VarDial shared task~\citep{2023-findings-vardial} and the addition of Bavarian German data from \citet{winkler-etal-2024-slot}. Dialectal NER data is drawn from WikiANN~\citep{pan-etal-2017-cross, rahimi-etal-2019-massively}.

Arabic sentiment analysis consists of multiple dialectal datasets aggregated within DialectBench~\citep{nabil-etal-2015-astd, 10.1007/978-3-319-60042-0_66, 8716394, 9068897, fourati-etal-2021-introducing, alqahtani-etal-2022-customer, Garouani_MAC, faisal-etal-2024-dialectbench}. Because the constituent datasets use different annotation schemes, we map all labels to a unified binary positive/negative classification task and exclude examples labeled as \textit{neutral} or \textit{objective}.

Arabic topic identification uses SIB-200~\citep{adelani2023sib200}, which is derived from FLORES~\citep{costa2022no}. Laurie Burchell (Senior Researcher at Common Crawl) has noted substantial overlap between Modern Standard Arabic (arb\_Arab) and three varieties in FLORES/SIB-200: Mesopotamian (acm\_Arab), Ta’izzi-Adeni (acq\_Arab), and Najdi Arabic (ars\_Arab).\footnote{\url{https://github.com/openlanguagedata/flores/issues/8}} We report full per-dialect results in Table~\ref{tab:full} as well as dialect averages excluding these varieties in Table~\ref{tab:arabic-overlap-sensitivity}; we find that the trends are the same, dialect average accuracy is consistently 1-2 points lower after removing the flagged varieties.

\paragraph{Evaluation} Recent work has shown that prompt-based evaluation of LLMs on dialectal tasks is sensitive to prompt design, instruction format, decoding behavior, and task structure \cite{faisal-anastasopoulos-2025-testing}. To isolate the effect of CPT on learned representations, we freeze all LLM parameters and train lightweight logistic regression classifiers on hidden representations. For sequence classification, we mean-pool final-layer token representations and report accuracy. For token classification, we use the first subword representation corresponding to each word and report macro F1.

\section{Results}

\begin{figure*}
    \centering
    \includegraphics[width=0.9\linewidth]{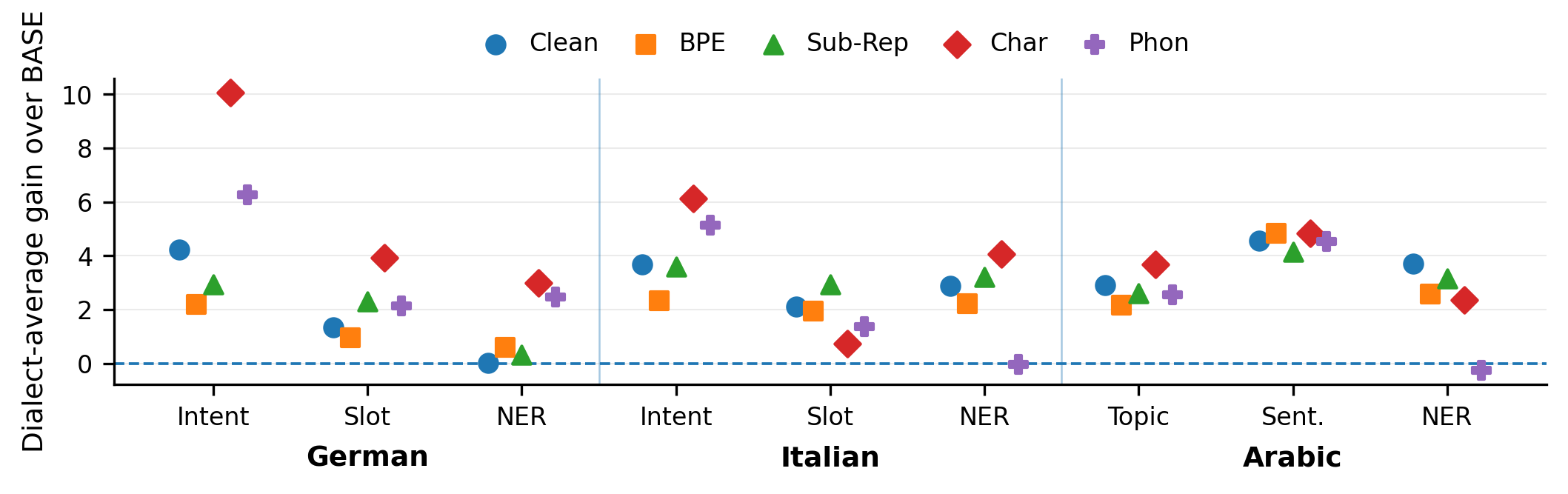}
    \caption{\textbf{Average improvement over \textsc{Base} across all language--task settings.} Continued pre-training consistently improves dialect performance, with \textsc{Char} producing the largest gains in most settings while leaving standard variety performance largely unchanged.}
    \label{fig:improvement}
\end{figure*}

\subsection{Perturbation-based CPT reliably improves dialect robustness.}

\label{sec:overall-results}

Table~\ref{tab:combined_main} reports dialect average performance, while Figure~\ref{fig:improvement} summarizes each CPT mode's improvement over \textsc{Base}. The central finding is straightforward: CPT consistently improves zero-shot dialect robustness. Across the 45 language--task--method dialect-average comparisons, 43 outperform \textsc{Base}, with an average gain of 3.0 points on dialect evaluation while preserving standard variety performance (0.79-point average improvement). These gains are also robust across random seeds. At the individual dialect-variety level, the 95\% confidence interval for improvement over \textsc{Base} lies entirely above zero in 116 of 160 comparisons, indicating that the observed improvements are not driven by a small number of favorable runs.

Having established that perturbation-based CPT reliably improves dialect robustness, we next examine how the different perturbation strategies compare and what accounts for their differing behavior.

\subsection{Character perturbations are the strongest general-purpose strategy but not a universal winner.}
\label{sec:method-tradeoffs}

Although all CPT modes tested improve dialect robustness, they do not do so equally. \textsc{Char} achieves the highest dialect-average performance in six of the nine language--task settings, making it the strongest overall strategy in our evaluation (Table~\ref{tab:combined_main}; Figure~\ref{fig:improvement}). Its advantage is particularly pronounced for German, where it consistently produces the largest improvements across tasks.

At the same time, no single perturbation strategy is uniformly optimal. \textsc{BPE} performs best on Arabic sentiment, \textsc{Clean} on Arabic NER, and \textsc{Sub-Rep} on Italian slot labeling. In several settings the differences among methods are also small, suggesting that the choice of perturbation is language- and task-dependent rather than universally fixed.

From a practical perspective, these results suggest that character perturbations provide the strongest default choice when no target-dialect supervision is available. At the same time, the relatively small differences among methods indicate that benchmark performance alone is insufficient to explain why some perturbations outperform others. We therefore turn to analyses of language modeling behavior and representation similarity to understand the mechanisms underlying these gains.

\paragraph{Qwen} To assess whether these conclusions generalize beyond a single model family, we repeat a subset of the experiments on Qwen2-7B\footnote{\url{https://huggingface.co/Qwen/Qwen2-7B}} using \textsc{Base}, \textsc{Clean}, and \textsc{Char} (Table~\ref{tab:qwen}). Although intentionally smaller in scope, the qualitative trends are preserved: \textsc{Char} again provides the strongest dialect performance while largely maintaining standard variety accuracy. We observe one exception in NER, where \textsc{Base} remains strongest on German and Arabic. Qwen2-7B was pre-trained on substantially more multilingual data than Llama-2-7B. This likely explains its stronger baseline performance on WikiAnn NER, leaving less room for further improvement. Overall, these experiments suggest that our principal conclusions are not specific to a single decoder-only LLM family.

\subsection{Similar downstream performance arises from distinct mechanisms.}
\label{sec:mechanistic-results}

The comparable downstream performance of several CPT strategies raises an important question: do they learn the same form of robustness, or arrive at similar performance through different mechanisms? To answer this, we compare two complementary views of adaptation. Bits per character (BPC) measures language model fit, with lower values indicating better prediction of natural text. Standard--dialect cosine similarity measures how similarly the model represents semantically equivalent standard and dialectal inputs, providing a proxy for invariance to surface-form variation. Table~\ref{tab:bpc} reports BPC, and Table~\ref{tab:cosine-appendix} reports cosine similarity.

\begin{figure}
    \centering
    \includegraphics[width=0.7\linewidth]{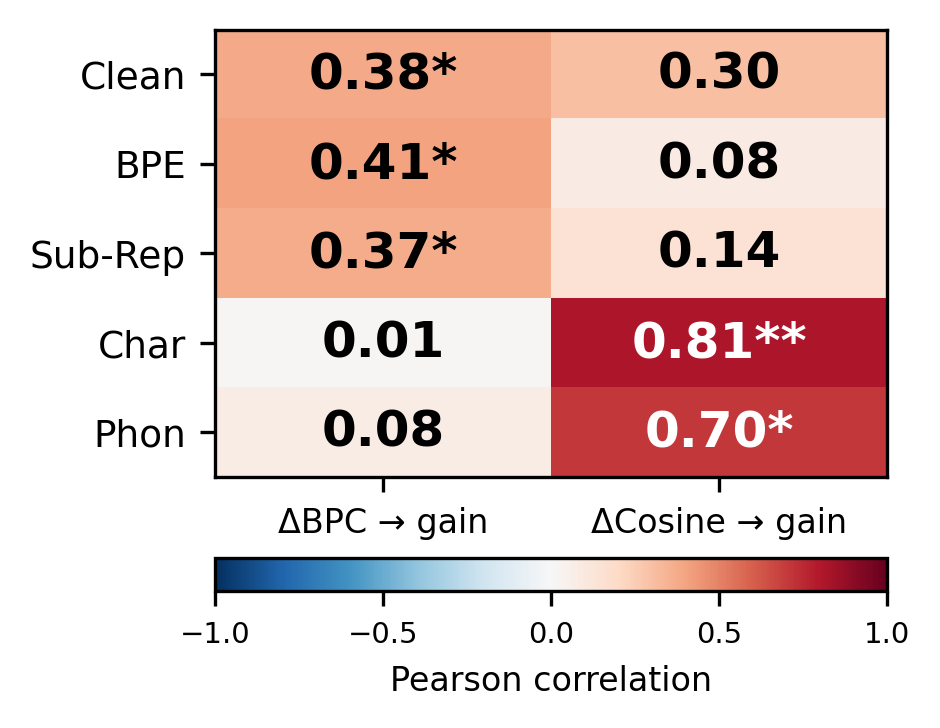}
    \caption{\textbf{Associations between mechanistic changes and downstream improvement.} Each cell reports the Pearson correlation between improvement over \textsc{Base} and either bits-per-character (BPC) improvement or increased standard--dialect representation similarity. Significant correlations are indicated by asterisk.}
    \label{fig:correlation_heatmap}
\end{figure}

The two analyses reveal different rankings. \textsc{Clean} achieves the largest improvement in language model fit, while \textsc{Char} produces the largest increase in standard--dialect representation similarity, followed closely by \textsc{Phon}. The contrast is particularly pronounced for German, while Arabic begins from a much higher baseline similarity, leaving less room for any increase.

More importantly, \textbf{Figure~\ref{fig:correlation_heatmap} shows that downstream improvements are associated with fundamentally different mechanisms across perturbation families.} For \textsc{Clean}, \textsc{BPE}, and \textsc{Sub-Rep}, improvements are significantly associated with better language model fit (Pearson's $r=0.376$, $p=.034$; $r=0.405$, $p=.021$; and $r=0.371$, $p=.037$, respectively), while correlations with representational similarity are weak and nonsignificant. The opposite pattern emerges for perturbations that directly modify surface forms. For \textsc{Char}, downstream improvement is strongly associated with increased standard--dialect representation similarity ($r=0.810$, $p=.003$) but essentially unrelated to BPC improvement ($r=0.012$, $p=.947$). \textsc{Phon} exhibits the same qualitative behavior, with a strong association between downstream gains and representational similarity ($r=0.703$, $p=.016$) but not language model fit ($r=0.077$, $p=.677$).

These analyses show that perturbation-based CPT does not induce a single form of robustness. Methods with comparable downstream performance arrive there through different behavioral and representational changes.

\subsection{CPT gains arise primarily from repairing existing failures.}
\label{sec:repair-results}

Aggregate improvements do not reveal whether CPT corrects existing errors or exchanges one set of mistakes for another. To distinguish these behaviors, we align predictions from \textsc{Base} and each CPT model and measure the proportion of \textsc{Base} errors that are corrected and the proportion of correct predictions by \textsc{Base} that become incorrect post-CPT. Their difference is the \emph{net repair rate}.

\begin{figure}
    \centering
    \includegraphics[width=\linewidth]{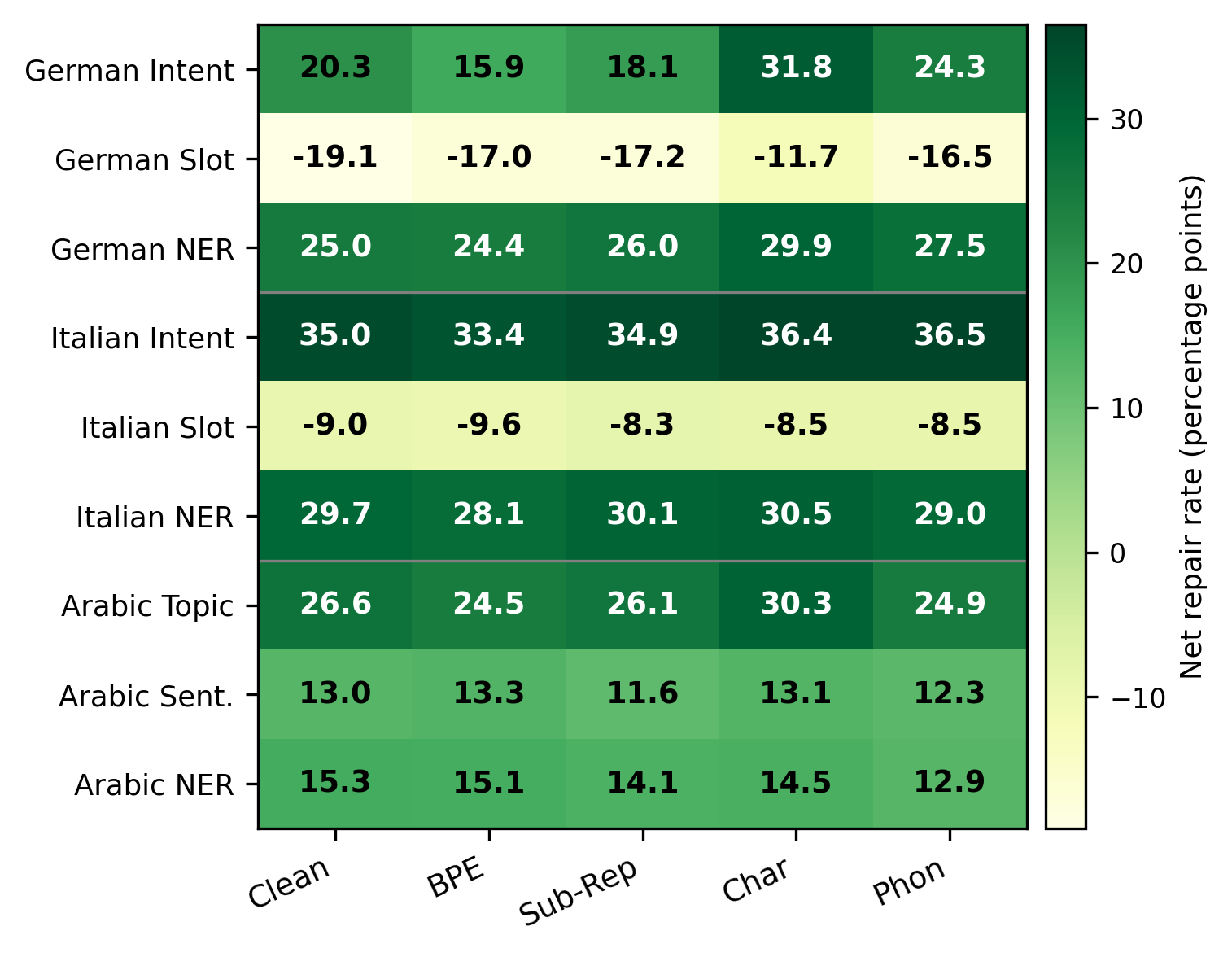}
    \caption{\textbf{Prediction transitions relative to \textsc{Base}.} Net repair rate is the difference between wrong to correct transitions (base to post-CPT) and correct to wrong. Across all settings except slot filling, all CPT modes repair substantially more errors than introduced.}
    \label{fig:transition}
\end{figure}

\begin{table}
\centering
\small
\begin{tabular}{p{0.97\columnwidth}}
\toprule

\textbf{German Intent Detection (South Tyrolean)}\\
\emph{stell an Wecker af 10 in der friah}\\
EN: Set an alarm for 10 in the morning.\\
Base: \textsc{bookrestaurant} \\
Char: \textsc{alarm/set\_alarm} \cmark\\

\midrule

\textbf{Italian Intent Detection (Neapolitan)}\\
\emph{Vota Soul Music cu 0}\\
EN: Rate ``Soul Music'' with a score of 0.\\
Base: \textsc{playmusic} \xmark \\
Char: \textsc{ratebook} \cmark\\

\midrule

\textbf{Arabic Topic Identification (Ta'izzi--Adeni Arabic)}\\
EN: Aerosmith cancelled the remaining concerts on their tour.\\
Base: \textsc{politics} \xmark \\
Char: \textsc{entertainment} \cmark\\

\bottomrule
\end{tabular}

\caption{\textbf{Prediction repair examples produced by \textsc{Char}.} In each example, \textsc{Base} predicts an incorrect label, while \textsc{Char} recovers the correct prediction. English translations are provided for readability.}
\label{tab:prediction_gallery}
\end{table}

Figure~\ref{fig:transition} shows that all CPT strategies improve robustness primarily by repairing existing errors rather than introducing new ones. Although the magnitude varies across perturbation methods, every strategy exhibits a strongly positive net repair profile, with \textsc{Char} producing the largest overall gains. This indicates that the improvements in Table~\ref{tab:combined_main} arise mainly from correcting \textsc{Base} failures.

The task-level heatmaps reveal where these improvements occur. German intent and NER exhibit particularly strong repair rates for \textsc{Char}, while Italian intent benefits consistently across all perturbation strategies. The notable exception is slot filling, where every CPT strategy exhibits a negative net repair rate. Here, changes to previously correct token predictions outweigh repaired errors, despite the overall improvements in slot filling performance reported in Table~\ref{tab:combined_main}. This apparent discrepancy reflects the different quantities being measured: transition analysis treats each token prediction equally, while macro-F1 depends on the overall balance of precision and recall across entity spans.

Table~\ref{tab:prediction_gallery} provides examples of prediction repairs made by \textsc{Char}. Across languages and tasks, the examples mirror the aggregate trend, illustrating that perturbation-based CPT primarily recovers errors made by the baseline model rather than introducing new prediction behavior.

\subsection{Cross-linguistic differences reveal where CPT has the most room to help.}
\label{sec:crosslingual-results}

\begin{figure}
    \centering
    \includegraphics[width=\linewidth]{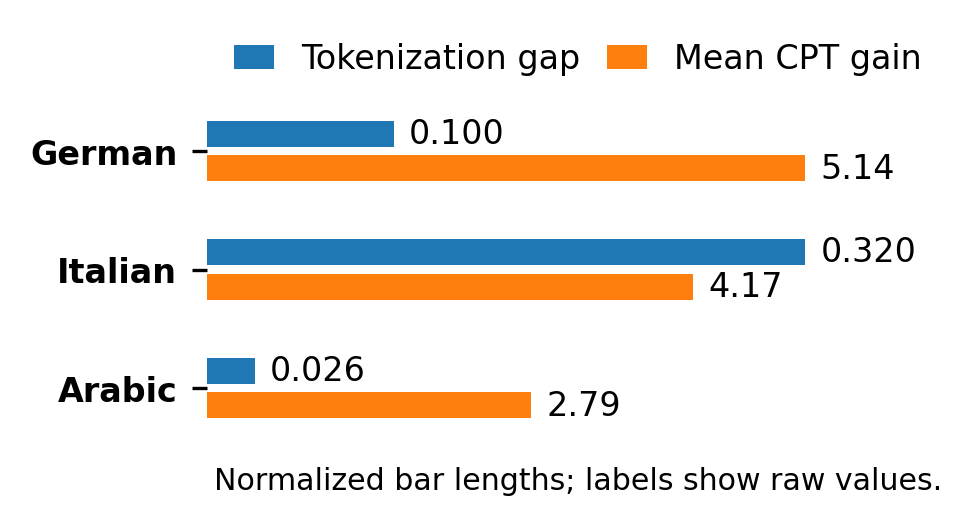}
    \caption{\textbf{Language-level relationship between tokenization mismatch and CPT gains.} Mean increase in tokens per word from standard to dialect text compared with the mean downstream improvement from CPT on the corresponding parallel-data task. German and Italian exhibit larger tokenization shifts and larger robustness gains than Arabic, suggesting that greater surface-form divergence provides more opportunity for perturbation-based adaptation.}
    \label{fig:tokenization}
\end{figure}

\begin{figure}
    \centering
    \includegraphics[width=\linewidth]{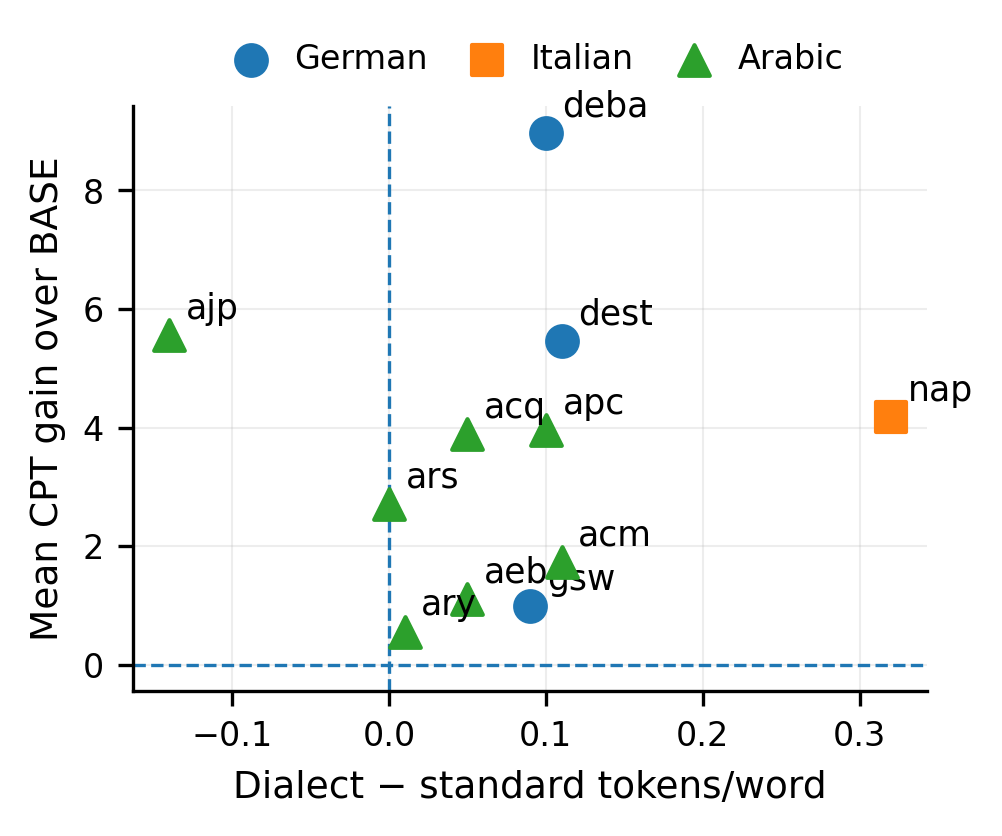}
    \caption{\textbf{Dialect-level relationship} between tokenization mismatch and CPT gains.}
    \label{fig:tokenization_per_variety}
\end{figure}

The magnitude of CPT improvements differs systematically across languages, motivating a closer look at how dialectal input differs from the corresponding standard variety. Figures~\ref{fig:tokenization} and~\ref{fig:tokenization_per_variety} compare tokenization differences between standard and dialect text with CPT  downstream gains. Paired with the tokenization statistics provided in Table~\ref{tab:tok_stats}, we see a clear language-level trend. Arabic dialects do not exhibit over-segmentation (more tokens/word, fewer characters/token) relative to the standard variety the way German and Italian dialects do. Accordingly, Arabic benefits less from perturbation-based CPT, while German and Italian exhibit larger gains. 

At the same time, Italian exhibits the largest increase in tokens per word, yet German receives the largest average improvement from CPT; these descriptive analyses therefore do not imply that tokenization differences cause robustness gains. Instead, they suggest that languages with greater surface-form divergence between standard and dialect varieties provide more opportunity for perturbation-based CPT to improve robustness, while the remaining variation likely reflects other lexical, morphological, and regional differences.

\subsection{Different strategies are effective in different linguistic conditions.}
\label{sec:method-interpretation}

Random character-level perturbations (\textsc{Char}) emerge as the strongest general-purpose strategy. They achieve the highest dialect performance in most language–task settings, produce the largest gains in standard–dialect representation similarity, and consistently attain the highest net repair rates. These advantages are particularly pronounced for German, whose dialects exhibit substantial orthographic variation and corresponding tokenization mismatch.  Phonologically informed perturbations (\textsc{Phon}) exhibit a similar pattern, consistently improving representation similarity while remaining competitive on downstream tasks. These findings suggest that exposing models to diverse surface forms during CPT promotes representations that remain stable despite spelling variation, without requiring explicit exposure to individual dialects.

In contrast, \textsc{Clean}, \textsc{BPE}, and \textsc{Sub-Rep} appear to improve robustness differently. These strategies yield larger improvements in language model fit (BPC) than in representation alignment, but these gains do not consistently translate into the strongest downstream performance. \textsc{Clean} remains competitive across nearly all settings, indicating that continued exposure to standard-language text improves the underlying language model, while explicit perturbations are often needed for the largest gains on dialects. The token-level strategies occupy an intermediate position: although \textsc{BPE} and \textsc{Sub-Rep} occasionally match or exceed \textsc{Char} on individual tasks (e.g., Arabic sentiment analysis, Italian slot filling), their results suggest that robustness to alternative token sequences alone is generally insufficient to maximize dialect robustness.

\section{Conclusion}

We presented a systematic study of perturbation-based continued pre-training (CPT) for improving zero-shot robustness to dialectal variation in multilingual decoder-only language models. Across German, Italian, and Arabic, our collection of CPT strategies consistently improved dialect robustness while largely preserving standard variety performance. Random character-level noise did best most often, although no single perturbation strategy was uniformly optimal across languages and tasks. Rather than constituting a single robustness technique, we find that perturbation-based CPT should be thought of as a family of adaptation strategies that achieve robustness through different mechanisms, with character-based perturbations more strongly promoting representational alignment and clean and token-level perturbations more strongly improving language model fit. We hope this perspective informs the design and evaluation of future robustness-oriented adaptation methods for multilingual language models.

\section*{Limitations}

Our evaluation primarily focuses on discriminative and representation-based tasks. While this setting isolates the effect of noisy CPT on learned representations and robustness to dialectal variation, it does not fully characterize behavior in open-ended generation or interactive settings. Future work should therefore evaluate instruction following, dialect-aware generation, and dialogue behavior across dialect communities.

\bibliography{custom}

\appendix
\label{sec:appendix}

\section{Supplemental Tables}

\paragraph{Full Results} We report performance for dialect averages in the main text. We provide results on all varieties (standard and dialects) individually across tasks in Table~\ref{tab:full}.

\begin{table*}[h]
\centering
\small
\setlength{\tabcolsep}{3.4pt}
\begin{tabular}{lllcccccc}
\toprule
Language & Task & Variety & BASE & CLEAN & BPE-DROP & SUB-REP & CHAR & PHON \\
\midrule
German & Intent & Standard German & 94.2 & \textbf{95.5 {$\pm$ 0.4}} & {95.2 {$\pm$ 0.3}} & 94.5 {$\pm$ 0.9} & \textbf{95.5 {$\pm$ 0.2}} & \textbf{95.5 {$\pm$ 0.3}} \\
 &  & Bavarian & 62.8 & 71.7 {$\pm$ 1.0} & 67.2 {$\pm$ 1.7} & 69.9 {$\pm$ 2.4} & \textbf{77.8 {$\pm$ 0.5}} & {72.3 {$\pm$ 0.8}} \\
 &  & South Tyrolean & 67.6 & 72.5 {$\pm$ 2.1} & 71.5 {$\pm$ 0.8} & 71.1 {$\pm$ 2.0} & \textbf{75.8 {$\pm$ 2.7}} & {74.4 {$\pm$ 0.5}} \\
 &  & Swiss German & 53.0 & 51.9 {$\pm$ 3.0} & 51.3 {$\pm$ 2.5} & 51.2 {$\pm$ 1.7} & \textbf{60.0 {$\pm$ 0.9}} & {55.5 {$\pm$ 0.8}} \\
German & Slot & Standard German & \textbf{30.8} & {28.7 {$\pm$ 0.7}} & 28.7 {$\pm$ 0.5} & 28.2 {$\pm$ 0.2} & 28.6 {$\pm$ 0.9} & 28.4 {$\pm$ 0.8} \\
 &  & Bavarian & 19.8 & 21.1 {$\pm$ 1.5} & 20.5 {$\pm$ 0.5} & 21.8 {$\pm$ 1.1} & \textbf{22.6 {$\pm$ 0.8}} & {22.2 {$\pm$ 0.7}} \\
 &  & South Tyrolean & 15.4 & 17.8 {$\pm$ 0.6} & 17.5 {$\pm$ 0.4} & {19.1 {$\pm$ 0.3}} & \textbf{20.7 {$\pm$ 0.5}} & 18.4 {$\pm$ 0.4} \\
 &  & Swiss German & 9.0 & 9.3 {$\pm$ 0.4} & 9.1 {$\pm$ 0.8} & {10.2 {$\pm$ 1.3}} & \textbf{12.7 {$\pm$ 0.9}} & 10.1 {$\pm$ 0.3} \\
German & NER & Standard German & 72.0 & 72.4 {$\pm$ 0.2} & {72.7 {$\pm$ 0.2}} & 72.6 {$\pm$ 0.2} & \textbf{72.7 {$\pm$ 0.1}} & 72.3 {$\pm$ 0.1} \\
 &  & Alsatian & 44.0 & 44.2 {$\pm$ 1.9} & 44.3 {$\pm$ 1.0} & 44.9 {$\pm$ 2.4} & \textbf{49.3 {$\pm$ 1.8}} & {46.1 {$\pm$ 1.9}} \\
 &  & Bavarian & 64.1 & 66.4 {$\pm$ 2.2} & \textbf{68.0 {$\pm$ 1.9}} & 65.1 {$\pm$ 0.3} & 65.2 {$\pm$ 0.6} & {67.8 {$\pm$ 2.4}} \\
 &  & K\"olsch & 31.6 & 32.3 {$\pm$ 1.3} & 32.1 {$\pm$ 1.5} & 31.9 {$\pm$ 1.1} & {35.9 {$\pm$ 0.6}} & \textbf{36.0 {$\pm$ 1.5}} \\
 &  & Luxembourgish & 38.8 & 38.7 {$\pm$ 1.3} & 39.7 {$\pm$ 0.7} & 39.6 {$\pm$ 1.9} & {42.9 {$\pm$ 0.8}} & \textbf{42.9 {$\pm$ 0.6}} \\
 &  & Limburgish & {43.8} & 41.0 {$\pm$ 1.4} & 42.3 {$\pm$ 1.5} & 42.6 {$\pm$ 1.2} & \textbf{46.1 {$\pm$ 0.7}} & 42.6 {$\pm$ 0.5} \\
 &  & Pennsylvania Dutch & 33.7 & 33.4 {$\pm$ 1.9} & 33.3 {$\pm$ 1.0} & 33.9 {$\pm$ 3.2} & {34.6 {$\pm$ 0.5}} & \textbf{35.5 {$\pm$ 0.7}} \\
\midrule
Italian & Intent & Standard Italian & 93.0 & \textbf{95.7 {$\pm$ 0.3}} & 95.6 {$\pm$ 0.3} & {95.7 {$\pm$ 0.5}} & 95.0 {$\pm$ 0.5} & 95.3 {$\pm$ 0.3} \\
 &  & Neapolitan & 70.4 & 74.1 {$\pm$ 0.6} & 72.7 {$\pm$ 2.3} & 74.0 {$\pm$ 1.0} & \textbf{76.5 {$\pm$ 2.2}} & {75.5 {$\pm$ 0.7}} \\
Italian & Slot & Standard Italian & 23.7 & {24.9 {$\pm$ 0.4}} & 24.3 {$\pm$ 0.2} & 24.8 {$\pm$ 1.2} & 24.7 {$\pm$ 1.1} & \textbf{25.3 {$\pm$ 1.1}} \\
 &  & Neapolitan & 18.1 & {20.2 {$\pm$ 1.5}} & 20.1 {$\pm$ 0.9} & \textbf{21.1 {$\pm$ 1.3}} & 18.9 {$\pm$ 0.3} & 19.5 {$\pm$ 0.5} \\
Italian & NER & Standard Italian & 76.2 & 76.6 {$\pm$ 0.1} & {76.7 {$\pm$ 0.1}} & \textbf{76.7 {$\pm$ 0.2}} & 76.7 {$\pm$ 0.2} & 76.5 {$\pm$ 0.2} \\
 &  & Neapolitan & 47.9 & {51.1 {$\pm$ 0.5}} & 49.3 {$\pm$ 1.4} & \textbf{51.8 {$\pm$ 1.3}} & 50.3 {$\pm$ 2.3} & 46.9 {$\pm$ 0.8} \\
 &  & Sicilian & 55.2 & 57.8 {$\pm$ 1.8} & {58.3 {$\pm$ 1.9}} & 57.7 {$\pm$ 0.3} & \textbf{61.0 {$\pm$ 0.8}} & 56.2 {$\pm$ 1.5} \\
\midrule
Arabic & Topic & MSA & 77.5 & {78.9 {$\pm$ 0.5}} & 77.3 {$\pm$ 0.6} & 78.3 {$\pm$ 1.2} & \textbf{79.4 {$\pm$ 1.7}} & 78.4 {$\pm$ 0.0} \\
 &  & Mesopotamian & 76.5 & \textbf{78.8 {$\pm$ 1.6}} & 77.5 {$\pm$ 0.8} & 77.9 {$\pm$ 0.5} & 78.3 {$\pm$ 1.0} & {78.6 {$\pm$ 0.7}} \\
 &  & Ta'izzi-Adeni & 75.5 & 79.4 {$\pm$ 0.0} & \textbf{79.7 {$\pm$ 0.7}} & 79.1 {$\pm$ 0.3} & {79.6 {$\pm$ 0.3}} & 79.1 {$\pm$ 0.7} \\
 &  & Tunisian & 72.5 & 73.7 {$\pm$ 1.0} & 71.6 {$\pm$ 2.2} & {74.2 {$\pm$ 2.0}} & \textbf{74.7 {$\pm$ 0.6}} & {74.2 {$\pm$ 2.0}} \\
 &  & South Levantine & 73.0 & {78.9 {$\pm$ 0.5}} & 78.4 {$\pm$ 0.8} & 77.5 {$\pm$ 0.5} & \textbf{80.6 {$\pm$ 2.0}} & 77.6 {$\pm$ 1.6} \\
 &  & North Levantine & 75.5 & \textbf{80.1 {$\pm$ 1.2}} & {79.4 {$\pm$ 1.0}} & 79.1 {$\pm$ 2.4} & \textbf{80.1 {$\pm$ 2.0}} & 78.6 {$\pm$ 1.0} \\
 &  & Najdi & 76.5 & \textbf{79.7 {$\pm$ 0.3}} & 78.6 {$\pm$ 1.0} & 79.1 {$\pm$ 0.6} & {79.6 {$\pm$ 1.2}} & 78.9 {$\pm$ 0.5} \\
 &  & Moroccan & 67.6 & 67.0 {$\pm$ 1.2} & 67.2 {$\pm$ 1.0} & {68.6 {$\pm$ 1.3}} & \textbf{70.3 {$\pm$ 0.7}} & 68.0 {$\pm$ 1.5} \\
Arabic & Sentiment & MSA & 81.7 & \textbf{83.6 {$\pm$ 0.3}} & 83.0 {$\pm$ 0.3} & 83.3 {$\pm$ 0.5} & 82.9 {$\pm$ 0.1} & {83.5 {$\pm$ 0.5}} \\
 &  & Tunisian & \textbf{53.9} & 52.3 {$\pm$ 0.6} & 52.3 {$\pm$ 1.1} & {52.5 {$\pm$ 0.8}} & 52.3 {$\pm$ 0.6} & 52.1 {$\pm$ 0.8} \\
 &  & Lebanese & 65.7 & {68.4 {$\pm$ 0.5}} & 67.9 {$\pm$ 0.6} & 67.3 {$\pm$ 0.4} & 68.0 {$\pm$ 0.9} & \textbf{69.5 {$\pm$ 1.2}} \\
 &  & Algerian & 41.1 & \textbf{50.1 {$\pm$ 3.2}} & {50.0 {$\pm$ 2.0}} & 48.8 {$\pm$ 0.9} & 50.0 {$\pm$ 5.3} & 49.5 {$\pm$ 3.3} \\
 &  & ary\_arab & 60.9 & 63.2 {$\pm$ 0.8} & 62.9 {$\pm$ 0.6} & 62.8 {$\pm$ 0.5} & \textbf{63.4 {$\pm$ 1.0}} & {63.3 {$\pm$ 0.9}} \\
 &  & Egyptian & 50.6 & {52.6 {$\pm$ 1.4}} & 52.3 {$\pm$ 1.5} & 51.4 {$\pm$ 0.1} & \textbf{53.0 {$\pm$ 2.1}} & 50.5 {$\pm$ 1.5} \\
 &  & Jordanian & 60.2 & 70.1 {$\pm$ 1.7} & {70.7 {$\pm$ 0.3}} & 70.4 {$\pm$ 1.3} & 70.2 {$\pm$ 1.1} & \textbf{71.5 {$\pm$ 1.1}} \\
 &  & Saudi & 49.3 & 56.8 {$\pm$ 4.5} & \textbf{59.3 {$\pm$ 3.1}} & 57.5 {$\pm$ 3.5} & {58.5 {$\pm$ 2.2}} & 57.1 {$\pm$ 2.2} \\
Arabic & NER & MSA & 64.1 & \textbf{65.6 {$\pm$ 0.1}} & {65.6 {$\pm$ 0.2}} & 65.1 {$\pm$ 0.4} & 65.3 {$\pm$ 0.3} & 64.9 {$\pm$ 0.4} \\
 &  & Egyptian & 50.3 & \textbf{54.0 {$\pm$ 1.2}} & 52.9 {$\pm$ 1.1} & {53.5 {$\pm$ 0.7}} & 52.7 {$\pm$ 1.3} & 50.1 {$\pm$ 0.1} \\
\bottomrule
\end{tabular}
\caption{Full per-variety results. CPT entries report mean and sample standard deviation across three random seeds; BASE is evaluated once. Best results (absolute) in each row are bold.}
\label{tab:full}
\end{table*}

\paragraph{Sensitivity to Flagged Arabic Dialects}
We present dialect average results for Arabic topic identification with and without the dialects flagged as being duplicates of MSA in Table~\ref{tab:arabic-overlap-sensitivity}. We find that the trends hold, the scores are simply about 1 to 2 points lower for each measurement.
\begin{table}[h]
\centering
\small
\setlength{\tabcolsep}{5pt}
\begin{tabular}{lcc}
\toprule
\textbf{Mode} &
\textbf{All varieties} &
\textbf{Excluding overlap} \\
\midrule
\textsc{Base}    & 73.88            & 72.18            \\
\textsc{Clean}   & 76.80 $\pm$ 0.58 & 74.92 $\pm$ 0.83 \\
\textsc{BPE}     & 76.05 $\pm$ 0.56 & 74.14 $\pm$ 0.74 \\
\textsc{Sub-Rep} & 76.49 $\pm$ 0.82 & 74.84 $\pm$ 1.23 \\
\textsc{Char}    & \textbf{77.57 $\pm$ 0.57} &
                   \textbf{76.39 $\pm$ 1.13} \\
\textsc{Phon}    & 76.42 $\pm$ 0.11 & 74.59 $\pm$ 0.37 \\
\bottomrule
\end{tabular}
\caption{\textbf{Sensitivity of Arabic topic-identification accuracy to varieties with reported standard variety overlap.}
The first column averages over all evaluated Arabic varieties. The second excludes Mesopotamian (\texttt{acm}), Ta'izzi--Adeni (\texttt{acq}), and Najdi Arabic (\texttt{ars}), which have been reported to contain substantial overlap or near-duplication with Modern Standard Arabic in FLORES/SIB-200. Values are mean accuracy and standard deviation across three CPT seeds. \textsc{Base} is evaluated once and therefore has no across-seed standard deviation.}
\label{tab:arabic-overlap-sensitivity}
\end{table}

\paragraph{Calibrated Parameters} Calibrated parameters for CPT modes inducing comparable token inflation rates (tokens/word increase after perturbation) are presented in Table~\ref{tab:calibration}.
\begin{table}[h]
\centering
\small
\setlength{\tabcolsep}{3pt}
\begin{tabular}{llccc}
\toprule
\textbf{CPT Mode} & \textbf{Lang} & \textbf{Sampling} & \textbf{Max} & \textbf{Observed} \\
\midrule
\textsc{BPE-Drop} & DE & $d \sim \mathcal{U}(0,d_{\max})$ & 0.23 & 0.06 \\
            & IT & $d \sim \mathcal{U}(0,d_{\max})$ & 0.24 & 0.06 \\
            & AR & $d \sim \mathcal{U}(0,d_{\max})$ & 0.24 & 0.06 \\
\midrule
\textsc{Sub-Rep} & All & $p = 0.05$ & 0.05 & 0.05 \\
\midrule
\textsc{Char} & All & $p \sim \mathcal{U}(0,1)$ & 1.00 & 0.25 \\
\midrule
\textsc{Phon} & DE & $\theta \sim \mathcal{U}(0,\theta_{\max})$ & 0.45 & 0.11 \\
                      & IT & $\theta \sim \mathcal{U}(0,\theta_{\max})$ & 0.30 & 0.07 \\
                      & AR & $\theta \sim \mathcal{U}(0,\theta_{\max})$ & 0.25 & 0.06 \\
\bottomrule
\end{tabular}
\caption{Perturbation rates used during CPT. \emph{Observed} reports the empirical average parameter value sampled over eligible documents during training. BPE dropout rate and DialUp phonological theta were calibrated separately for each language to approximately match the realized perturbation level of \textsc{Char}.}
\label{tab:calibration}
\end{table}

\paragraph{CPT Hyperparameters}
\label{sec:hyperparams}
All CPT conditions use the same optimization configuration. Hyperparameters are summarized in Table~\ref{tab:cpt_hyperparameters}, chosen on the basis of \citet{kojima2025continual}. All reported results for Llama average three independent CPT runs. Quantization is done using \texttt{bitsandbytes} \cite{dettmers2023qlora}. Models are evaluated using the 2500-step checkpoint, as training loss consistently plateaued after this point. Each CPT run requires approximately 30 hours on a single NVIDIA RTX A6000 GPU.
\begin{table}[h]
\centering
\small
\begin{tabular}{ll}
\toprule
Parameter & Value \\
\midrule
Base model & Llama-2-7B \\
PEFT & LoRA ($r=16$, $\alpha=16$) \\
Optimizer & AdamW \\
Learning rate & $2\times10^{-4}$ \\
$\beta_1,\beta_2$ & 0.9, 0.95 \\
Weight decay & 0.01 \\
Warmup & 200 steps \\
LR schedule & Cosine \\
Sequence length & 1024 \\
Effective batch size & 32 \\
Quantization & 4-bit \\
Training steps & 2500 \\
\bottomrule
\end{tabular}
\caption{\textbf{Hyperparameters} for CPT experiments.}
\label{tab:cpt_hyperparameters}
\end{table}

\paragraph{Qwen Supplemental Results}
We present results on a subset of experiments on Qwen2-7b in Table~\ref{tab:qwen} and find similar trends as our main experiments with Llama.
\begin{table}[h]
\centering
\small
\setlength{\tabcolsep}{5pt}
\begin{tabular}{llccc}
\toprule
\textbf{Language} & \textbf{Setting} &
\textbf{Base} &
\textbf{Clean} &
\textbf{Char} \\
\midrule

\multirow{2}{*}{German Intent}
& Standard & 95.00 & 95.00 & 95.00 \\
& Dialect  & 61.87 & 63.53 & \textbf{70.27} \\
\addlinespace

\multirow{2}{*}{German Slot}
& Standard & 39.54 & 39.36 & \textbf{40.78} \\
& Dialect  & 25.60 & 25.77 & \textbf{26.82} \\
\addlinespace

\multirow{2}{*}{German NER}
& Standard & 81.86 & 82.08 & \textbf{82.26} \\
& Dialect  & \textbf{63.62} & 61.87 & 62.41 \\
\addlinespace

\multirow{2}{*}{Arabic Topic}
& Standard & \textbf{85.29} & \textbf{85.29} & 84.80 \\
& Dialect  & 82.35 & 81.44 & \textbf{82.56} \\
\addlinespace

\multirow{2}{*}{Arabic Sentiment}
& Standard & \textbf{83.54} & 83.46 & 82.99 \\
& Dialect  & 62.17 & 61.22 & \textbf{67.88} \\
\addlinespace

\multirow{2}{*}{Arabic NER}
& Standard & \textbf{81.53} & 74.10 & 81.09 \\
& Dialect  & \textbf{66.35} & 53.60 & 66.21 \\

\bottomrule
\end{tabular}

\caption{\textbf{Evaluation on Qwen.}
We repeat a subset of the experiments using Qwen with \textsc{Base}, \textsc{Clean}, and character-level (\textsc{Char}) continued pre-training. Standard (S) and dialect (D) results are reported for the evaluated tasks. Although more limited than the primary experiments (single seed and fewer CPT conditions), the overall pattern of larger dialect improvements under character-level perturbation is largely preserved.}

\label{tab:qwen}
\end{table}

\paragraph{Datasets} Language varieties and tasks are presented in Table~\ref{tab:tasks}. Dataset statistics are presented in Table~\ref{tab:dataset_stats}.

\begin{table}[h]
\centering
\small
\setlength{\tabcolsep}{2pt}
\begin{tabular}{lccccc}
\toprule
\textbf{Dialect} &
\textbf{Intent} &
\textbf{Slot} &
\textbf{Topic} &
\textbf{Sentiment} &
\textbf{NER} \\
\midrule

\multicolumn{6}{l}{\textbf{German}} \\
\midrule
Standard German & \checkmark & \checkmark & -- & -- & \checkmark \\
Bavarian & \checkmark & \checkmark & -- & -- & \checkmark \\
South Tyrolean & \checkmark & \checkmark & -- & -- & -- \\
Swiss German & \checkmark & \checkmark & -- & -- & -- \\
Alsatian & -- & -- & -- & -- & \checkmark \\
Limburgish & -- & -- & -- & -- & \checkmark \\
Luxembourgish & -- & -- & -- & -- & \checkmark \\
Kölsch & -- & -- & -- & -- & \checkmark \\
Pennsylvania Dutch & -- & -- & -- & -- & \checkmark \\

\midrule
\multicolumn{6}{l}{\textbf{Italian}} \\
\midrule
Standard Italian & \checkmark & \checkmark & -- & -- & \checkmark \\
Neapolitan & \checkmark & \checkmark & -- & -- & \checkmark \\
Sicilian & -- & -- & -- & -- & \checkmark \\

\midrule
\multicolumn{6}{l}{\textbf{Arabic}} \\
\midrule
MSA & -- & -- & \checkmark & \checkmark & \checkmark \\
Mesopotamian & -- & -- & \checkmark & -- & -- \\
Ta’izzi-Adeni & -- & -- & \checkmark & -- & -- \\
Tunisian & -- & -- & \checkmark & \checkmark & -- \\
South Levantine & -- & -- & \checkmark & -- & -- \\
North Levantine & -- & -- & \checkmark & -- & -- \\
Najdi & -- & -- & \checkmark & -- & -- \\
Moroccan & -- & -- & \checkmark & \checkmark & -- \\
Egyptian & -- & -- & \checkmark & \checkmark & \checkmark \\
Lebanese & -- & -- & -- & \checkmark & -- \\
Algerian & -- & -- & -- & \checkmark & -- \\
Jordanian & -- & -- & -- & \checkmark & -- \\
Saudi & -- & -- & -- & \checkmark & -- \\

\bottomrule
\end{tabular}
\caption{\textbf{Language varieties and downstream tasks.} All classifiers are trained on standard language data and evaluated zero-shot on dialect test sets.}
\label{tab:tasks}
\end{table}

\begin{table}[h]
\small
\centering
\setlength{\tabcolsep}{3pt}
\begin{tabular}{llll}
\toprule
\textbf{Task} & \textbf{Source} & \textbf{Train Size} & \textbf{Test Size} \\
\midrule
Intent & SID4LR & 43k & 500 \\
Slot & SID4LR & 43k & 500 \\
Topic & SIB-200 & 701 & 204 \\
Sentiment & DialectBench & 8k & 0.5-15k \\
NER & WikiANN & 20k & 100 \\
\bottomrule
\end{tabular}
\caption{\textbf{Dataset statistics for downstream tasks.} Training set sizes correspond to the standard variety only; test set sizes are per dialect. Training data is used only for fitting linear probes.}
\label{tab:dataset_stats}
\end{table}

\paragraph{Language Abbreviations} Language abbreviations used in the main text are presented in Table~\ref{tab:language_codes}.
\begin{table}[h]
\centering
\small
\setlength{\tabcolsep}{8pt}
\begin{tabular}{ll}
\toprule
\textbf{Code} & \textbf{Language / Variety} \\
\midrule

\multicolumn{2}{l}{\textbf{German}}\\
de   & German\\
deba & Bavarian\\
dest & South Tyrolean\\
gsw  & Swiss German\\
\addlinespace

\multicolumn{2}{l}{\textbf{Italian}}\\
it   & Italian\\
nap  & Neapolitan\\
\addlinespace

\multicolumn{2}{l}{\textbf{Arabic}}\\
ar   & Modern Standard Arabic\\
acm  & Mesopotamian Arabic\\
acq  & Ta'izzi--Adeni Arabic\\
aeb  & Tunisian Arabic\\
ajp  & South Levantine Arabic\\
apc  & North Levantine Arabic\\
ars  & Najdi Arabic\\
ary  & Moroccan Arabic\\

\bottomrule
\end{tabular}
\caption{Language and dialect \textbf{abbreviations} used throughout the paper. Standard varieties are listed first within each language, followed by dialect varieties in alphabetical order.}
\label{tab:language_codes}
\end{table}

\paragraph{Bits per Character}
Table~\ref{tab:bpc} reports bits per character (BPC) averaged across seeds for each CPT mode. Lower BPC indicates better language modeling fit.
\begin{table}
\centering
\small
\setlength{\tabcolsep}{2pt}
\begin{tabular}{@{}lSSSSSS@{}}
\toprule
\textbf{Lang} & \textbf{Base} & \textbf{Clean} & \textbf{BPE-Drop} & \textbf{Sub-Rep}       & \textbf{Char} & \textbf{Phon}        \\ \midrule
\texttt{DE-S}               & 2.08 & \bf 1.75  & 2.18              & 2.06                   & 1.77 & 1.775 \\
              \texttt{DE-D} & 3.58          & 3.28           & 3.68              & 3.59                   & \bf 3.18 & 3.336          \\
\texttt{IT-S}               & 2.14          & \bf 1.76  & 2.176              & 2.079                   & 1.80 & 1.799                   \\
\texttt{IT-D}                & 3.33          & \bf 2.94  & 3.352              & 3.26                   & 3.02 & 3.080                   \\
\texttt{AR-S}             & 2.314          & \bf 1.746  & 1.861              & 1.889                   & 1.795 & 1.827                   \\
             \texttt{AR-D}          & 2.658 & \bf 2.176  & 2.298              & 2.313 & 2.21 & 2.244                   \\ \bottomrule
\end{tabular}
\caption{\textbf{Bits per character (BPC)} on standard-variety (S) and dialect-average (D) evaluation text. Lower values indicate better language modeling fit. Dialectal inputs generally exhibit higher BPC than standard-variety inputs under the base model. \textsc{Clean} achieves the lowest BPC across most settings, with \textsc{Noisy} consistently close behind. Results are shown for standard-dialect parallel datasets, German and Italian intent detection and Arabic topic detection.}
\label{tab:bpc}
\end{table}

\paragraph{Cosine Similarity}
Table~\ref{tab:cosine-appendix} reports cosine similarity between standard--dialect parallel sentences averaged across seeds for each CPT mode.

\begin{table}[t]
\centering
\small
\setlength{\tabcolsep}{2pt}
\begin{tabular}{lrrrrrr}
\toprule
\textbf{Lang} & \textbf{Base} & \textbf{Clean} & \textbf{BPE-Drop} & \textbf{Sub-Rep} & \textbf{Char} & \textbf{Phon} \\
\midrule
DE & 0.783 & 0.852 & 0.824 & 0.871 & 0.902 & 0.878 \\
IT & 0.858 & 0.886 & 0.871 & 0.903 & 0.930 & 0.911 \\
AR & 0.989 & 0.992 & 0.992 & 0.990 & 0.994 & 0.993 \\
\bottomrule
\end{tabular}
\caption{\textbf{Standard--dialect representational cosine similarity on parallel datasets.} Higher values indicate greater similarity between representations of aligned standard and dialectal inputs. Values are macro-averaged over dialect varieties and averaged across random seeds for CPT conditions; \textsc{Base} is evaluated once. Results are shown for standard-dialect parallel datasets, German and Italian intent detection and Arabic topic detection.}
\label{tab:cosine-appendix}
\end{table}

\end{document}